%% file: main.tex
\documentclass[10pt,twocolumn,letterpaper]{article}

\usepackage[pagenumbers]{cvpr} 

\input{preamble}

\definecolor{cvprblue}{rgb}{0.21,0.49,0.74}
\usepackage[pagebackref,breaklinks,colorlinks,citecolor=cvprblue]{hyperref}

\def\paperID{XXX} 
\def\confName{CVPR}
\def\confYear{2024}

\title{\model: A Mamba-Based Encoder-Decoder Architecture with a Statistical Verb-Noun Interaction Module for Video Action Forecasting @ Ego4D Long-Term Action Anticipation Challenge 2024}

\author{
Zeyun Zhong$^{1,2}$ \quad
Manuel Martin$^2$\quad 
Frederik Diederichs$^2$\quad
Juergen Beyerer$^{1,2}$ \vspace{1mm} \\
$^1$Karlsruhe Institute of Technology \qquad $^2$Fraunhofer IOSB \\ 
{\tt\small zeyun.zhong@kit.edu, \{manuel.martin, frederik.diederichs, juergen.beyerer\}@iosb.fraunhofer.de }
}

\comm{
\author{First Author\\
Institution1\\
Institution1 address\\
{\tt\small firstauthor@i1.org}
\and
Second Author\\
Institution2\\
First line of institution2 address\\
{\tt\small secondauthor@i2.org}
}
}

\begin{document}
\maketitle

\begin{abstract}
This report presents a novel Mamba-based encoder-decoder architecture, \model, featuring an integrated verb-noun interaction module that utilizes a statistical verb-noun co-occurrence matrix to enhance video action forecasting. This architecture not only predicts verbs and nouns likely to occur based on historical data but also considers their joint occurrence to improve forecast accuracy. The efficacy of this approach is substantiated by experimental results, with the method achieving second place in the Ego4D LTA challenge and ranking first in noun prediction accuracy.
\end{abstract}

\section{Introduction}
Action forecasting in video sequences is a critical task in many applications such as surveillance, human-computer interaction, and autonomous driving. Accurate prediction of future actions requires not only recognizing patterns in visual data but also understanding the context in which these actions occur.
Traditional methods have utilized recurrent neural networks (RNNs) such as GRUs \cite{farhaWhenWillYou2018} and LSTMs \cite{furnari2019egocentric}, along with transformers \cite{girdharAnticipativeVideoTransformer2021}, focusing on extended temporal contexts to deepen the understanding of past interactions and achieve progress in this field.

Recently, Mamba~\cite{gu2023mamba} has been introduced and has demonstrated superior capabilities in efficient long-sequence modeling. We adopt this model as our encoder to process long-term visual observations. For future action anticipation, we employ a query-based transformer decoder, following~\cite{gongFutureTransformerLongterm2022,zhong2023diffant}.

In the Ego4D \cite{graumanEgo4DWorld0002022} Long-Term Action Anticipation (LTA) challenge, participants are tasked with predicting a sequence of 20 future actions, described by verb-noun pairs, from video clips approximately five minutes long. Traditional approaches, which often predict verbs and nouns independently, may yield suboptimal results due to the mismatch in the affordances of objects and actions. To tackle this, we introduce a statistical verb-noun interaction module that utilizes a co-occurrence matrix to enhance prediction accuracy.
The details of our approach are presented in Section~\ref{sec:method}, followed by an experimental analysis in Section~\ref{sec:results}. We conclude the report with a discussion of our findings and limitations in Section~\ref{sec:conclusion}.

\begin{figure}[t]
    \centering
    \includegraphics[width=\linewidth]{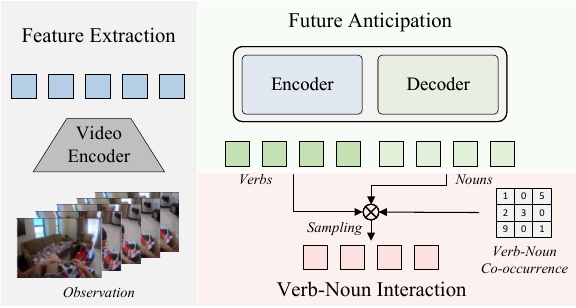}
    \caption{Schematic illustration of the proposed action forecasting architecture, featuring a dual-stage process with feature extraction via a video backbone and future anticipation using an encoder-decoder structure. The verb-noun interaction module leverages co-occurrence matrices to enhance predictive accuracy during inference by integrating statistical relationships between actions (verbs) and objects (nouns).}
    \label{fig:figure1}
\end{figure}

\begin{figure*}[t]
    \centering
    \begin{subfigure}{0.7\linewidth}
        \centering
        \includegraphics[width=\linewidth]{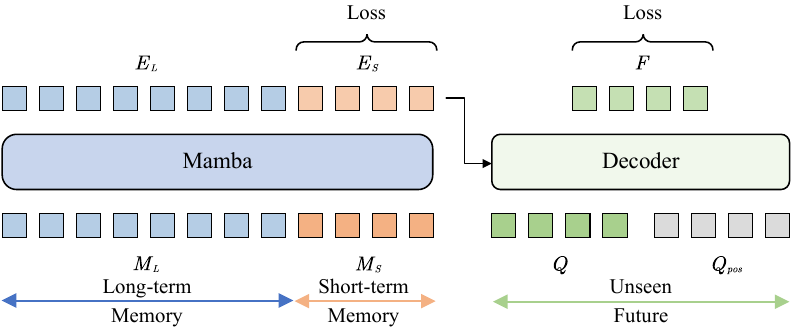}
        \caption{Architecture of the Mamba-based encoder-decoder model.}
    \end{subfigure} \hfill
    \begin{subfigure}{0.225\linewidth}
        \centering
        \includegraphics[width=\linewidth]{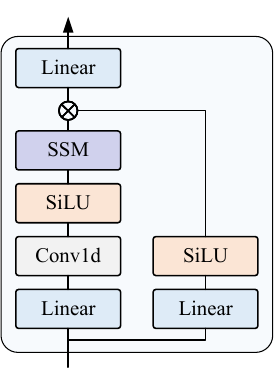}
        \caption{A Mamba block.}
        \label{fig:mamba}
    \end{subfigure}
    \caption{(a) The proposed model, \model, integrates long-term and short-term memories processed through the Mamba encoder. The future actions are anticipated by the decoder, utilizing static content embeddings $Q$ and learnable positional embeddings $Q_{pos}$, under the guidance of encoded past short-term memories $E_S$. (b) A Mamba block consists of multiple layers including linear layers, 1D convolutional layers, and SiLU (Sigmoid Linear Unit) activation functions, with the core being the SSM module.}
    \label{fig:architecture}
\end{figure*}

\section{Method}
\label{sec:method}

The proposed architecture is designed for video action forecasting, incorporating an encoder-decoder framework to efficiently process and predict future actions from video sequences. The architecture operates in two primary stages: feature extraction and future anticipation (Fig.~\ref{fig:figure1} and~\ref{fig:architecture}). Initially, video frames undergo feature extraction using a video backbone, e.g., VideoMAE~\cite{tong2022videomae}, which compresses the visual information into a sequence of visual embeddings. This extracted data is then passed through the Mamba encoder~\cite{gu2023mamba} to incorporate long-range temporal context, making them task-oriented (Sec~\ref{sec:encoder}). The output from the encoder feeds into the decoder, which anticipates potential future actions using learnable queries (Sec~\ref{sec:decoder}). Finally, a statistical verb-noun interaction module is utilized to improve the sampled predictions of the decoder during inference (Sec~\ref{sec:verb-noun}).

\subsection{Preliminaries}
In this work, we choose Mamba~\cite{gu2023mamba}, a state space model (SSM), as our encoder due to its superior and efficient long-sequence modeling ability. Below, we give a short introduction to the SSM-based models.

The SSM-based models are inspired by the continuous system, which maps a 1-D function or sequence $x(t) \in \mathbb{R} \mapsto y(t) \in \mathbb{R}$ through a hidden state $h(t) \in \mathbb{R}^\mathtt{N}$. This system uses $\mathbf{A} \in \mathbb{R}^{\mathtt{N} \times \mathtt{N}}$ as the evolution parameter and $\mathbf{B} \in \mathbb{R}^{\mathtt{N} \times 1}$, $\mathbf{C} \in \mathbb{R}^{1 \times \mathtt{N}}$ as the projection parameters.
\begin{equation}
\begin{aligned}
\label{eq:lti}
h'(t) &= \mathbf{A}h(t) + \mathbf{B}x(t), \\
y(t) &= \mathbf{C}h(t).
\end{aligned}
\end{equation}

The discrete version of the continuous system includes a timescale parameter $\mathbf{\Delta}$ to transform the continuous parameters $\mathbf{A}$, $\mathbf{B}$ to discrete parameters $\mathbf{\overline{A}}$, $\mathbf{\overline{B}}$. The commonly used method for transformation is zero-order hold (ZOH), which is defined as follows:

\begin{equation}
\begin{aligned}
\label{eq:zoh}
\mathbf{\overline{A}} &= \exp{(\mathbf{\Delta}\mathbf{A})}, \\
\mathbf{\overline{B}} &= (\mathbf{\Delta} \mathbf{A})^{-1}(\exp{(\mathbf{\Delta} \mathbf{A})} - \mathbf{I}) \cdot \mathbf{\Delta} \mathbf{B}.
\end{aligned}
\end{equation}

After the discretization of $\mathbf{\overline{A}}$, $\mathbf{\overline{B}}$, the discretized version of Eq.~\eqref{eq:lti} using a step size $\mathbf{\Delta}$ can be rewritten as:
\begin{equation}
\begin{aligned}
\label{eq:discrete_lti}
h_t &= \mathbf{\overline{A}}h_{t-1} + \mathbf{\overline{B}}x_{t}, \\
y_t &= \mathbf{C}h_t.
\end{aligned}
\end{equation}

\subsection{Mamba Encoder}
\label{sec:encoder}

The parameters $\mathbf{A}$, $\mathbf{B}$, and $\mathbf{C}$ in traditional SSM models are time-invariant, meaning that they are independent of the input. To incorporate context-aware ability (selectivity), Mamba~\cite{gu2023mamba} introduced additional linear layers to make these parameters dependent on the input.
An overview of a Mamba block is illustrated in Fig.~\ref{fig:mamba}, which contains multiple layers including Linear transformations, 1D Convolutional layers, and SiLU (Sigmoid Linear Unit) activation functions, with the core being the selective SSM module.

Given a full observation $M \in \mathbb{R}^{(L+S) \times D'} $, where $(L+S)$ denotes the total observation length and $D'$ represents the hidden dimension of the extracted visual embeddings, we first use a linear layer to adjust the hidden dimension from $D'$ to the model's hidden dimension $D$. We then split it into long-term memory $M_L \in \mathbb{R}^{L \times D}$ and short-term memories $M_S \in \mathbb{R}^{S \times D}$. While the long-term memory provides additional temporal context, we calculate the loss only on the short-term memory, following~\cite{wang2023memory,an2023miniroad}. 
 After processing by the Mamba encoder, the long-term $M_L$ and short-term memories $M_S$ are converted to $E_L \in \mathbb{R}^{L \times D}$ and $E_S \in \mathbb{R}^{S \times D}$, respectively. 

\subsection{Query Decoder}
\label{sec:decoder}

The decoder follows the query-based paradigm~\cite{carion2020end,zhu2020deformabledetr}, originally proposed to eliminate handcrafted components in object detection and recently applied in long-term action anticipation~\cite{gongFutureTransformerLongterm2022,zhong2023diffant}. It takes queries in form of static content embeddings $Q \in \mathbb{R}^{Z \times D}$ (often implemented as zero vectors) and learnable positional embeddings $Q_{pos} \in \mathbb{R}^{Z \times D}$ as input, generating future embeddings $F \in \mathbb{R}^{Z \times D}$, under the guidance of encoded past observations $E_S$ via cross-attention. $Z$ represents the number of future actions to be predicted, i.e., 20 for Ego4D long-term action anticipation task.
The temporal order of queries is aligned with the sequence of future actions, \ie, each query directly corresponds to a respective future action, following previous work~\cite{gongFutureTransformerLongterm2022,zhong2023diffant}.

\subsection{Verb-Noun Interaction}
\label{sec:verb-noun}

Given the predicted future embeddings $F$, we apply two separate classification heads to generate probabilities for verbs $F_{Verbs} \in \mathbb{R}^{Z \times V}$ and nouns $F_{Nouns} \in \mathbb{R}^{Z \times N}$, and we calculate losses on these predictions using standard cross-entropy loss. $V$ and $N$ represent the number of verb and noun classes, respectively. During inference, we sample possible future actions from the predicted verbs and nouns. 
However, we note that sampling verbs and nouns independently and then simply combining them to represent a potential future action is sub-optimal for two main reasons. Firstly, objects (nouns) have their own affordances, meaning some interactions may not be permissible for specific objects, for instance, a book is readable but not eatable. Secondly, each dataset has its own statistics, while some combinations of verbs and nouns may occur frequently in our daily lives, they may not necessarily appear in the dataset. To address this, we calculate all co-occurrences of verb-noun combinations in Ego4D and use this information as a prior to improve the sampling accuracy during inference. Specifically, we first calculate the joint probabilities of predicted verbs and nouns $F_{Actions} \in \mathbb{R}^{Z \times V \times N}$ by multiplying $F_{Verbs}$ and $F_{Nouns}$ using tensor broadcasting. We then multiply $F_{Actions}$ with the normalized co-occurrence matrix $O \in \mathbb{R} ^{V\times N}$ to obtain dataset-aware joint probabilities, from which the final verbs and nouns are jointly sampled. 

\section{Experiments}
\label{sec:results}
\subsection{Implementation Details}
\myparagraph{Feature extraction.} We use VideoMAE~\cite{tong2022videomae} for feature extract, which was pre-trained on Kinetics~\cite{carreiraQuoVadisAction2017} and then finetuned on Ego4D by~\cite{chen2022internvideo}. We employ a sliding window approach with a window size of 16 frames, and feed 16 consecutive frames totalling 0.533s of video at a frame rate of 30fps to the model to generate one visual embedding. Since \cite{chen2022internvideo} has finetuned two models for verb and noun classification respectively, we concatenate the embeddings from these two models and feed the concatenated embeddings into our proposed model. 

\myparagraph{Model Details.} We use four layers in both the Mamba encoder and decoder. The hidden dimension is set to 1024. Our model observes 64 seconds for long-term memory and 30 seconds for short-term memory. For training, the batch size is set at 128, and the learning rate is set at 0.0001.

\myparagraph{Evaluation metric.} We evaluate our method using the metric of edit distance. Given $K$ predicted sequences with each consisting $Z$ future actions $\left\{\left(\hat{n}_z^k, \hat{v}_z^k\right)\right\}_{z=1}^Z{}_{k=1}^K$, the action edit distance is measured as the best individual edit distance with the ground truth action sequence $\left\{(n_z, v_z)\right\}_{z=1}^Z$,
\[
\Delta_{\text{tot}} = \min_k \Delta_e\left(\left\{\left(\hat{n}_z^k, \hat{v}_z^k\right)\right\}_{z=1}^Z, \left\{(n_z, v_z)\right\}_{z=1}^Z\right).
\]
Similar metrics can be defined for verb and noun predictions respectively.

\subsection{Analysis}
\begin{table}[t]
    \centering
    \begin{tabular}{cccccc}
    \toprule
       \multicolumn{3}{c}{Loss}  & \multirow{2}{*}{Verb $\downarrow$} & \multirow{2}{*}{Noun $\downarrow$} & \multirow{2}{*}{Action $\downarrow$}  \\
       \cmidrule(lr){1-3}
       verb & noun & action & \\
    \midrule
       \checkmark  & \checkmark & \checkmark & 0.9622 & 0.7362  & 0.9551 \\
       \checkmark & \checkmark & \xmark & 0.7721 & 0.6733 & 0.9242\\
    \bottomrule
    \end{tabular}
    \caption{Edit distance on the validation set (V2). Results for actions are based on our defined action taxonomy. Including an action classification loss term in the training objective proves to be harmful.}
    \label{tab:validation_results}
\end{table}

Since we calculate the co-occurrence of all verb-noun pairs in the dataset, we generate an action taxonomy containing all verb-noun pairs that have at least one sample in the dataset. This results in a total of 7328 actions. With such a taxonomy, we can directly predict actions and provide the model with additional supervision beyond just verb and noun predictions. To evaluate the effectiveness of our introduced action taxonomy, we conduct experiments, where the model predicts either verbs, nouns, and actions, or only verbs and nouns. In the case where actions are directly predicted (first row of Table~\ref{tab:validation_results}), we show the edit distance of the predicted actions, along with the verbs and nouns derived from the actions using the taxonomy. When only verbs and nouns are predicted, we convert these pairs to actions using our defined taxonomy. To enable a deterministic comparison, results in Table~\ref{tab:validation_results} are based on the classes with maximal predicted probabilities. Our introduced verb-noun interaction module, described in~\ref{sec:verb-noun}, is therefore not applied here.

As observed in Table~\ref{tab:validation_results}, the inclusion of the action loss term in the training objective is harmful. This could be attributed to the extreme class imbalance in the dataset, as many of the 7328 actions have only one sample. Consequently, we remove the action loss term in our final model.

\begin{table}[t]
\centering
\setlength\tabcolsep{4.5pt}
\begin{tabular}{clccc}
\toprule
Rank & Participant team & Verb $\downarrow$ & Noun $\downarrow$ & Action $\downarrow$ \\ 
\midrule
1 & EgoVideo (newdict) & 0.6354 & 0.6367 & 0.8504 \\
\rowcolor{Gray} 2 & HAI-PUI & 0.6362 & 0.6258 & 0.8663 \\ 
3 & Charlie's (AntGPT) & 0.6531 & 0.6446 & 0.8748 \\ 
4 & live (live1+) & 0.6885 & 0.6714 & 0.8840 \\ 
5 & Doggeee & 0.6956 & 0.6506 & 0.8856 \\ 
6 & PaMsEgoAI & 0.6838 & 0.6785 & 0.8933 \\ 
7 & Brunos & 0.7004 & 0.7092 & 0.9142 \\ 
8 & Slowfast V2 & 0.7169 & 0.7359 & 0.9253 \\ 
\bottomrule
\end{tabular}
\caption{Challenge results on the test set. Results marked in gray indicate our second-place ranking on the challenge leaderboard.}
\label{tab:test_results}
\end{table}

\subsection{Challenge Results}

The results of the leading teams on the current leaderboard are shown in Table~\ref{tab:test_results}. With the proposed method, our team, ``HAI-PUI'', highlighted in gray, ranks second overall. Notably, our model achieves the best edit distance of 0.6258 for noun predictions among all entries.

\section{Conclusion and Limitations}
\label{sec:conclusion}
Our Mamba-based encoder-decoder architecture with a statistical verb-noun interaction module has demonstrated impressive results in the Ego4D long-term action anticipation challenge. Our team ranks second overall, excelling in noun prediction with the lowest edit distance score. We note that our model is trained exclusively on Ego4D LTA data, which may lack commonsense knowledge and may not generalize well to unseen data. We believe that combining our model with zero-shot large language models (LLMs) would yield better results. Additionally, the proposed statistical verb-noun interaction in its current form is only applied during inference. We hypothesis that making the model aware of the verb-noun co-occurrences would further improve performance.

{
    \small
    \bibliographystyle{ieeenat_fullname}
    \bibliography{main}
}


\end{document}

%% file: preamble.tex
\usepackage[dvipsnames]{xcolor}
\definecolor{airforceblue}{rgb}{0.36, 0.54, 0.66}

\newcommand{\comm}[1]{\iffalse #1 \fi}
\usepackage{colortbl}
\usepackage{soul}
\usepackage{pifont}
\usepackage{lipsum}
\usepackage{hhline}
\usepackage{booktabs}
\usepackage{subcaption}
\usepackage{multirow}
\usepackage{multicol}
\usepackage{xspace}
\usepackage{float}

\newcommand{\xmark}{\ding{55}}%
\definecolor{Gray}{gray}{0.9}
\newcommand{\model}{QueryMamba\xspace}
\newcommand{\myparagraph}[1]{\vspace{3.5pt}\noindent\textbf{#1}}